\newif\ifarxiv
\arxivfalse

\newif\ifdraft
\draftfalse

\documentclass[eat,twocolumn]{jmlr}

\usepackage{multirow}
\usepackage{chngcntr}

\usepackage[normalem]{ulem} 
\newcommand{\stkout}[1]{\ifmmode\text{\sout{\ensuremath{#1}}}\else\sout{#1}\fi}
\ifdraft
\newcommand{\added}[1]{\textcolor{blue}{#1}}
\newcommand{\deleted}[1]{\textcolor{red}{\stkout{#1}}}
\newcommand{\deletedcitep}[1]{{\color{red}#1}}

\newcommand{\deletedfloat}[1]{}
\newcommand{\commented}[1]{\textcolor{blue}{#1}}
\else
\newcommand{\added}[1]{#1}
\newcommand{\deleted}[1]{}
\newcommand{\deletedcitep}[1]{}

\newcommand{\deletedfloat}[1]{}
\newcommand{\commented}[1]{}
\fi

\usepackage{longtable}

\usepackage{booktabs}
\usepackage[load-configurations=version-1]{siunitx} 

\theorembodyfont{\upshape}
\theoremheaderfont{\scshape}
\theorempostheader{:}
\theoremsep{\newline}

\jmlrvolume{}
\firstpageno{1}

\jmlryear{2022}
\jmlrworkshop{Machine Learning for Health (ML4H) 2022}

\title[Structured State Space Models for ECGs]{Advancing the State-of-the-Art for ECG Analysis through Structured State Space Models}

\author{
	\Name{Temesgen Mehari} \Email{temesgen.mehari@ptb.de}\\
	\addr Physikalisch-Technische Bundesanstalt, Berlin, Germany\\
	\addr Fraunhofer Heinrich-Hertz-Institut, Berlin, Germany
	\AND
	\Name{Nils Strodthoff} \Email{nils.strodthoff@uni-oldenburg.de} \\
	\addr Carl von Ossietzky Universität Oldenburg, Oldenburg, Germany
}

\begin{document}

\maketitle

\begin{abstract}
	The field of deep-learning-based ECG analysis has been largely dominated by convolutional architectures. This work explores the prospects of applying the recently introduced structured state space models (SSMs) as a particularly promising approach due to its ability to capture long-term dependencies in time series. We demonstrate that this approach leads to significant improvements over the current state-of-the-art for ECG classification, which we trace back to individual pathologies. Furthermore, the model's ability to capture long-term dependencies allows to shed light on long-standing questions in the literature such as the optimal sampling rate or window size to train classification models. Interestingly, we find no evidence for using data sampled at 500Hz as opposed to 100Hz and no advantages from extending the model's input size beyond 3s. Based on this very promising first assessment, SSMs could develop into a new modeling paradigm for ECG analysis.

\end{abstract}

\section{Introduction}
Machine learning and deep learning in particular deep learning has the potential to transform the entire field of healthcare. The electrocardiogram (ECG) is particularly suited to spearhead this development due to its widespread use (in the US during about 5\% of the office visits, an ECG was ordered or provided \citep{NACMS2016}). While it only requires basic recording equipment, it holds enormous diagnostic potential that we only gradually start to uncover assisted by machine learning \citep{Hannun2019,Attia2019,Lima2021,Verbrugge2022}. On the algorithmic side, the analysis of ECGs based on raw sensor data is still largely dominated by convolutional neural networks \citep{strodthoff2020deep, hannun2019cardiologist,Attia2019,Ribeiro2020}. This default choice is slowly challenged by the rise of transformer-based architectures or combinations of convolutional architectures with attention elements as exemplified by the winning solutions of the two past editions of the Computing in Cardiology Challenge \citep{Natajan2020Wide,Nejedly2021}. In this work, we explore a novel algorithmic approach, \textit{Structured State Space Sequence (S4) Models} \citep{gu2021efficiently}, which learns a continuous representation of time series data and are particularly suited for modeling long-term dependencies. \\
We put the model through a thorough evaluation using \added{an} established benchmarking procedure \citep{strodthoff2020deep} on the \textit{PTB-XL} \citep{Wagner:2020PTBXL,Wagner2020:ptbxlphysionet,Goldberger2020:physionet} and \textit{Chapman} \citep{Zheng2020} datasets demonstrating consistent improvements over the existing (mostly convolutional) state-of-the-art methods on both datasets, which we trace back to improvements in the predictive performance for individual ECG statements. \\
We demonstrate the model's ability for evaluation at sampling rates unseen during training, essentially without any loss in predictive performance. Furthermore, we use the model capability to capture long-term dependencies to systematically investigate several long-standing questions in the field, i.e.\, how long-ranged are interactions in ECG data that have to be captured explicitly by the model and do models actually profit from input data at a sampling frequency of 500~Hz as compared to 100~Hz. 

\section{Related Work}
\noindent\textbf{ECG classification} The field of ECG analysis is largely dominated by convolutional architectures, see \citep{Hong2020, petmezas2022state} for a recent reviews. The superiority of modern ResNet- or Inception-based convolutional architectures was confirmed in an extensive comparative study on the \textit{PTB-XL}  dataset \citep{strodthoff2020deep}. This is in line with the excellent performance of such architectures on a broad range of time series classification tasks, see \citep{IsmailFawaz2020inceptionTime}. Interestingly, this supremacy was already challenged in \citep{Mehari:2021Self}, where the convolutional baseline was outperformed by a large recurrent neural network with convolutional feature extractor. Therefore, it represents a natural question to ask whether architectures that are even more adapted to the necessities of time series can lead to further performance improvements.\\
\noindent\textbf{Structured State Space Models for clinical time series} The motivation for the development of structured state space models (SSMs) was the wish to devise an architecture that is suited to capture long-term dependencies in very long temporal data, including medical time series as a particular example \citep{NEURIPS2021_05546b0e}. To support the applicability to the latter, the authors considered a classical vital sign prediction task on ECG and PPG time series as input \citep{NEURIPS2021_05546b0e,Gu2022hippo2} and clearly outperformed the current state-of-the-art for this tasks. These results represent a very encouraging sign for the application of these models in the broader context of medical time series. Nevertheless the prior study cannot be considered as a comprehensive ECG analysis task, which is the topic of this work.
In a different line of work, SSMs were used to model the internal state in diffusion models for time series imputation \citep{juan2022}, which lead to unprecedented imputation quality (among others) for ECG data, which provides additional hints for the potential advantages of SSMs also in a purely discriminative setting. We therefore aim to investigate the motivating claims for SSMs in the context of ECG data.
\section{Methods}
\subsection{Models}
\noindent\textbf{Structured State Space Models}
Structured State Space Models (SSMs) were introduced in \citep{gu2021efficiently} showing outstanding results on problems that require capturing long-range dependencies. The model consists of stacked \textit{S4 layers} that in turn draw on state-space models, frequently used in control theory, of the form
\begin{equation}
\begin{split}
	x'(t)=&\textbf{\textit{A}}x(t) + \textbf{\textit{B}}u(t)\,, \\
	y(t)=&\textbf{\textit{C}}x(t) + \textbf{\textit{D}}u(t)\,,
\end{split}
\label{eq:state_space}
\end{equation}
that map a one-dimensional input $u(t)\in\mathbb{R}$ to a one-dimensional output $y(t)\in \mathbb{R}$ mediated through a hidden state $x(t)\in {\mathbb{R}}^N$ parametrized through matrices $A,B,C,D$. These continuous-time parameters can be mapped to discrete-time parameters $\bar{A}, \bar{B}, \bar{C}$ for a given step size $\Delta$.
These allow to form the \textit{SSM convolutional kernel} $\bar{K}(\bar A, \bar B, \bar C)$ that allows to calculate the output $y$ by a simple convolution operation, $y = \bar{K} \ast u$. One of the main contribution of \citep{gu2021efficiently} lies in providing a stable and efficient way to evaluate the kernel $\bar{K}$. Second, building on earlier work \citep{NEURIPS2020_102f0bb6}, they identify a particular way, according to HiPPO theory \citep{NEURIPS2020_102f0bb6}, of initializing the matrix $A\in {\mathbb{R}}^{n\times n}$ as key to capture long-range interactions. $H$ copies of such layers parameterizing a mapping from $\mathbb{R}\to\mathbb{R}$ are now concatenated and fused through a point-wise linear operation to form a S4 layer mapping from ${\mathbb{R}}^H \to {\mathbb{R}}^H$.\\
\noindent \textbf{Supervised model} The model used for supervised training follows the original S4 architecture \citep{gu2021efficiently} and consists of a convolutional layer as input encoder, followed by four \textit{S4 blocks} which are connected through residual connections interleaved with normalization layers, with a global pooling layer and a linear classifier on top. The S4 blocks comprise the S4 layer accompanied by dropout and GeLU activations and a linear layer, see \citep{gu2021efficiently} for details. The architecture is summarized schematically in Fig.~\ref{fig:model}. We use the shorthand notation \textit{4S4+1FC} to refer to this model. 
\subsection{Datasets and experimental procedure} To benchmark the performance of the proposed approach, we use two large, publicly available 12-lead ECG datasets, namely the \textit{PTB-XL} \citep{Wagner:2020PTBXL,Wagner2020:ptbxlphysionet,Goldberger2020:physionet} and the \textit{Chapman} \citep{Zheng2020} datasets. For \textit{PTB-XL}, we follow the benchmarking methodology established in \citep{strodthoff2020deep} and report the performance in terms of macro AUC (on the test fold) on the most finegrained level with 71 labels covering a broad range of diagnostic, form-related and rhythm-related ECG statements. For \textit{Chapman}, we use the rhythm statements provided as primary annotations in the dataset. We mimic the procedure followed on \textit{PTB-XL}  as closely as possible by forming 10 stratified folds (8 training, 1 validation, 1 test fold) and also use macro AUC as performance metric. To ensure that each fold contains at least one sample of each ECG statement, we filter out statements that occur fewer than ten times. This reduces the label set from 68 to 48 statements.\\
For the experiments involving S4 layers, we use a batch size of 32, $N=8$ for the state dimension in the S4-Layers (as optimal value identified on the \textit{PTB-XL}  validation set) and train the model with a constant learning rate schedule and a learning rate $lr=0.001$ for 50 epochs with the AdamW Optimizer \citep{adamw2017}. If not stated otherwise, we train models on input sequences of length 2.5s in order to remain comparability to results in the literature. During training, subsequences are randomly cropped from the input record. During test time, we use test-time-augmentation: We create ten overlapping crops of the same length as the input size used during training from the original record, using different strides such that the whole sample gets covered, and take the mean of their respective output probabilities as final prediction for the entire sample. 
\section{Results}

\noindent \textbf{SSMs outperform the current supervised state-of-the-art}
As mentioned above, state-of-the-art approaches in deep-learning-based ECG analysis mainly rely on modern convolutional architectures. As a representative example, we use a model with \textit{xresnet1d50} architecture that was shown to lead to competitive results compared to the state-of-the-art at that time \citep{Mehari:2021Self}. We also compare to a recurrent neural network with convolutional feature extractor (\textit{4FC+2LSTM+2FC}) \citep{Mehari:2021Self} that showed the best reported supervised performance on \textit{PTB-XL} to date. In Table~\ref{tab:supervised}, we compare these models to the \textit{4S4+1FC} model based on comprehensive ECG classification tasks on the \textit{PTB-XL} and \textit{Chapman} datasets. On \textit{PTB-XL}, the  \textit{4S4+1FC} model outperforms all baseline methods in a statistical significant manner, see below for a detailed description. Interestingly, the ranking of the algorithms is largely consistent on the \textit{Chapman} dataset but the differences between the different approaches are not significant in this case. This might be explained by the fact that all models achieve very high predictive performance on this dataset with macro AUC values beyond 0.98, which is not the ideal situation if one aims to quantify performance differences in a statistically significant manner. \\
For a proper discussion of the results, we have to refine our notion of statistical and systematical uncertainty measures. Following the methodology used in  \citep{Mehari:2021Self}, we consider two sources of uncertainty, the statistical uncertainty due to the randomness of the training process, which can be assessed through multiple training runs, and the uncertainty due to the finiteness and the specificity of the label distribution of the test set. We address the latter via empirical bootstrapping ($n_\text{iter}=1000$ iterations) on the test set. Comparing two particular trained models, we consider the performance difference to be statistically significant if the bootstrap 95\% confidence intervals for the performance difference does not overlap with zero. We address the uncertainty due to the stochasticity of the training process by training $n_\text{runs}=10$ for each of the models we aim to compare. For each of the $n_\text{runs}^2$ comparisons, we assess the statistical significance via bootstrapping as defined above. Finally, we define a model to perform statistically significantly better/worse in case that 60\% of the model comparisons turn out to be statistically significantly better/worse. Just like the significance level, this threshold can be chosen at will as long as it exceeds 40\% for consistency reasons \citep{Springenberg2022Histo} and relates to the amount of uncertainty due to fluctuations across training runs one is willing to tolerate. \\
Furthermore, in Appendix \ref{app:paths}, we look into specific pathologies to investigate whether or to what extent the improved performance of the \textit{4S4+1FC} model can be traced back to certain pathologies.  Summarizing, we find statistical significant improvements for 8 (SARRH, PAC, ISCIN,STD\_, ABQRS, SR, IMI, NORM) of the 71 ECG statements, while only for 2AVB the performance decreased consistently.

\begin{table*}[ht]
	\centering
	
	\caption{Comparing supervised performance of the state-of-the-art models on two large ECG datasets. Here and in the following, we report mean and standard deviation of the test set scores over 10 runs using a concise error notation where e.g.\ 0.9175(39) signifies $0.9175\pm 0.0039$. The asteriks stands for statistically significant better performance than \textit{4FC+2LSTM+2FC} (as the previous state-of-the-art).}
	\label{tab:supervised}
	\begin{tabular}{l|l|l}
		\toprule
		&\multicolumn{2}{c}{dataset}\\
		& \textit{PTB-XL} &  \textit{Chapman}\\
		
		\midrule
		
		\textit{xresnet1d50} \citep{strodthoff2020deep} & 0.9286(28) &  0.9805(39) \\
		\textit{4FC+2LSTM+2FC} (causal) \citep{Mehari:2021Self} &0.9295(31) &0.9854(12) \\\hline
		\textit{4S4+1FC} (this work)& \textbf{0.9417(16)*} &  \textbf{0.9876(11)}\\
		
		\bottomrule	
		
	\end{tabular}
	
\end{table*}


\noindent \textbf{SSMs allow inference at unseen sampling rates}
A compelling aspect of state space models is that, due to the continuous character of the state-transition matrix $\textbf{\textit{A}}$ in equation \ref{eq:state_space}, the model can be evaluated on data that was sampled at a different rate than the training data, by simply adjusting the step size in the discretization step during test time. Table \ref{tab:model_selection} depicts a cross-evaluation matrix, in which we trained and cross-evaluated the \textit{4S4+1FC} model on 100, 200 and 500Hz. We see no or just minor losses in performance when varying test from train sampling rates, even if the sampling rates differ by a factor of 5. This is a particular asset of SSM models as it avoids the necessity to resample the data, which might otherwise be a source of additional systematic uncertainties. 
\ifarxiv
\else
\begin{table*}[ht]
	\centering
	\caption{
		Comparing supervised performance of the \textit{4S4+1FC} model trained/tested on different sampling rates.}
	\label{tab:model_selection}
	\begin{tabular}{ll|l|l|l}
		\toprule
		&&\multicolumn{3}{c}{test}\\
		&& 100Hz & 200Hz & 500Hz\\
		
		\midrule
		
		\multirow{3}{*}{\rotatebox[origin=c]{90}{train}}
		&100Hz & 0.9417(35) & 0.9418(36) & 0.9416(36)\\
		
		&200Hz    & 0.9414(16)  & 0.9416(16)& 0.9417(16)\\
		
		&500Hz   & 0.9415(18) & 0.9421(19)& 0.9421(19) \\
		\bottomrule	
	\end{tabular}
	
\end{table*}
\fi

\begin{figure}[ht]
	\centering
	\includegraphics[width=0.5\textwidth]{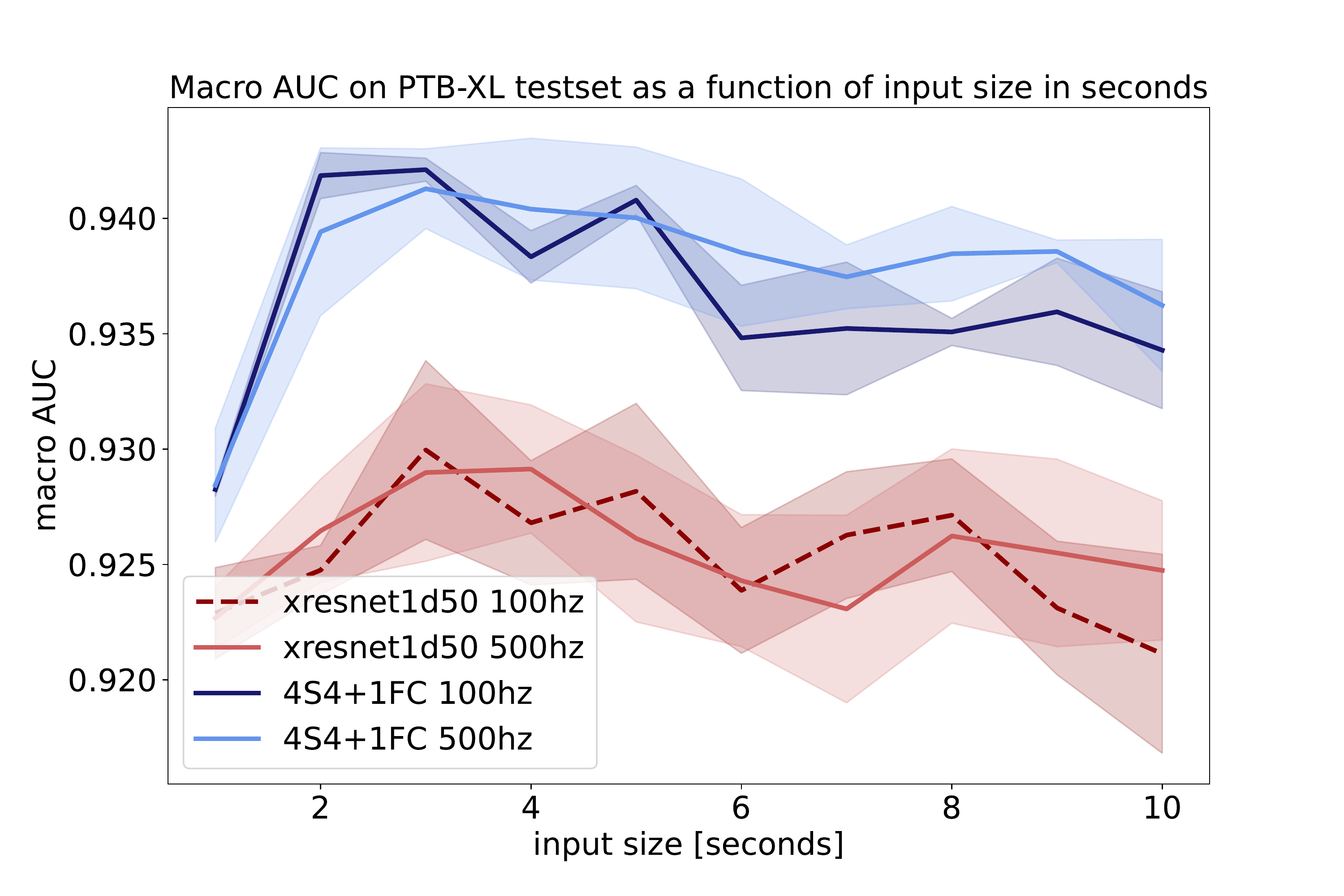}
	\label{fig:input_size}
	\caption{Model performance in dependence of (physical) input size for two convolutional models (\textit{xresnet1d50}) at 100/500Hz as compared to two SSMs (\text{4S4+1FC}) at 100/500Hz.}
\end{figure}

\noindent \textbf{No long-term dependencies in ECG data beyond 3s}
In this paragraph, we aim to clarify in a quantitative way, how long-ranged interactions are actually present in ECG data, which is a long-standing question that has not been systematically addressed so far. We address this question in terms of the size of the input window that is passed to the model (while performing test-time-augmentation, i.e.\, combining information from different segments for the sample-level prediction at all times). We believe that this question could not be answered in an unbiased way so far due to the inability of prior architectures to capture long-term dependencies in the data without adjusting hyperparameters such as kernel sizes etc.\ in the case of convolutional models.
In Fig.~\ref{fig:input_size}, we investigate the model performance (macro AUC measured on the \textit{PTB-XL} test set) as a function of input size for two convolutional models using input data sampled at 100Hz and 500Hz and two SSMs using the same kind of input data. As described above, we train models on various input sizes, measured in physical units for comparability, and obtain aggregated predictions for the full samples during test time by taking the average of ten input windows that are consequently moved through the signal with varying stride. As first observation from Fig.~\ref{fig:input_size}, we see that the performance gap between convolutional models and structured state space models, which was already apparent at an input size of 2.5s in Tab.~\ref{tab:supervised}, persists across all input sizes. Second, within each model architecture, the results from input data sampled at 100Hz as compared to 500Hz largely overlap. This already puts into question the potential advantage of using input data at 500Hz for ECG classification purposes. We will revisit this question in a more detail in the next paragraph. Third, the plot shows an interesting dependence on the input size that is qualitatively consistent across model architectures: The performance from aggregated model predictions shows a peak around input sizes around 2-3s. This hints at the fact that for ECG classification tasks based on short 12-lead ECG data, the ability of the model to explicitly capture long-range interactions beyond about 3s is not beneficial. On the contrary, the models seems to profit more from averaging overlapping predictions from different sliding windows. This observation is very much in line with the fact that most pathologies affect all beats equally (with a few exceptions such as premature ventricular contractions) and the fact that for average heart rates between 60 and 100bpm, a sliding window of 3s already contains 3-5 beats. This question is obviously completely independent from the question of capturing long-term dependencies in long-term ECGs with potentially rapid rhythm changes within the sample and should be revisited in this context in future work.

\noindent\textbf{No significant advantages from sampling frequencies beyond 100Hz}
We also revisit the question of the comparison between sampling frequencies at 100Hz vs. 500Hz in a statistically more rigorous manner based on the methodology presented above. At a input size of 2.5s and for fixed model architecture, we find no statistically significant performance difference between both sampling frequencies. This applies equally well to the level of individual label AUCs, in an analysis analogous to the one carried out for the comparison between 4S4+1FC and xresnet above. We want to stress that this statement obviously strongly depends on the label distribution of the dataset under consideration in the sense that there might be systematic improvements for certain ECG statements that do not turn out to be statistically significant due to large fluctuations as a consequence of small samples sizes.


\section{Summary}
In this work, we used structured state space models, which are particularly suited to capture long-term dependencies in time series, to challenge the supremacy of convolutional architectures in the field of deep-learning-based ECG analysis. In fact, we were able to demonstrate consistent improvements over the previous state-of-the-art on large, comprehensive ECG classification datasets\deleted{ both in the supervised and in the self-supervised regime}. We use the model's ability to capture long-term dependencies to shed new light on the question of the optimal sampling frequency and the model's input size. The code underlying our experiments is publicly available under \url{https://github.com/tmehari/ssm_ecg}.

\acks{This project (18HLT07 MedalCare) has received funding from the EMPIR programme co-financed by the Participating States and from the European Union’s Horizon 2020 research and innovation programme.
}

\bibliography{s4_extended_abstract_ml4h}

\appendix
\counterwithin{figure}{section}
\counterwithin{table}{section}
\setcounter{figure}{0}
\setcounter{table}{0}
\section{Datasets \& Models}
\label{app:datmod}
\begin{figure}
	\centering
	\includegraphics[width=0.5\textwidth, angle=270]{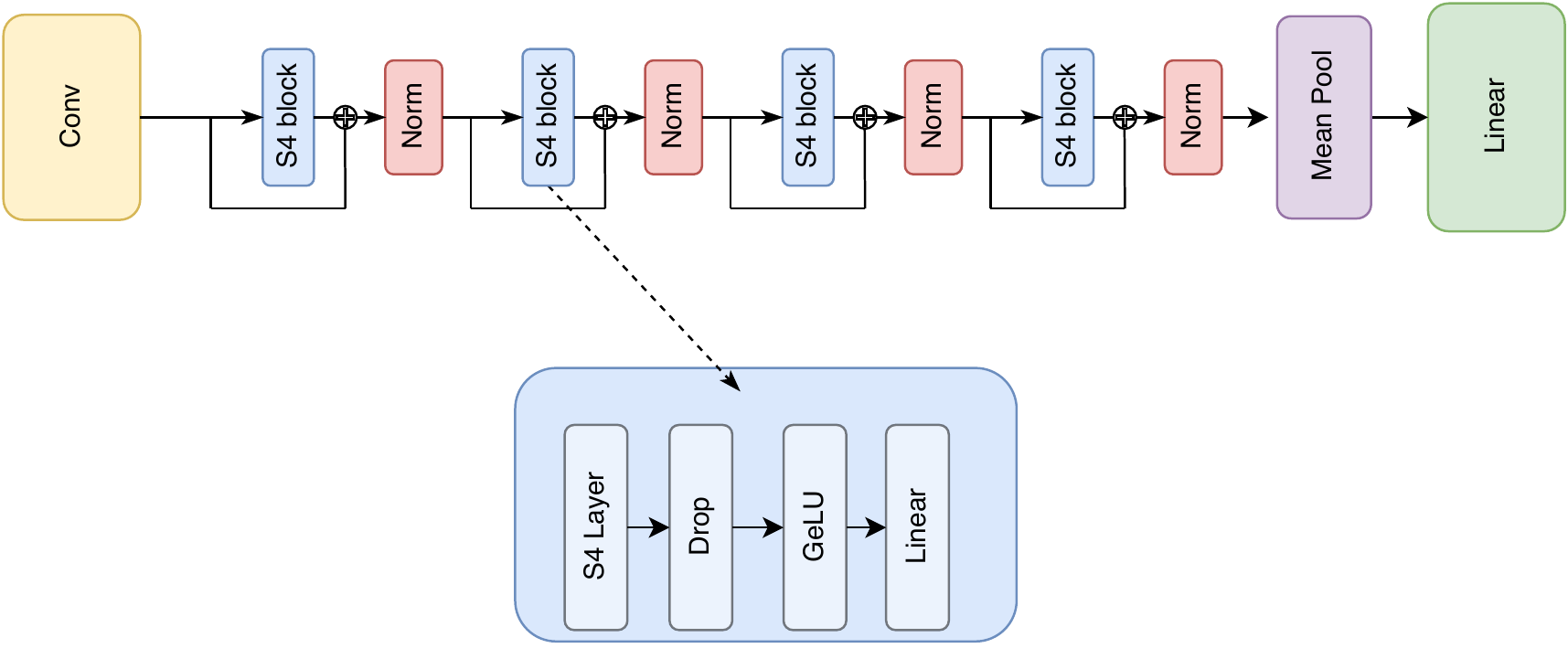}
	\caption{The model used during supervised training follows the original S4 architecture \citep{gu2021efficiently}. The model consists of a convolutional layer as input encoder, followed by four S4 blocks which are connected through residual connections with a normalization layer. The prediction is obtained from a linear layer following a mean pooling layer.
	}
	\label{fig:model}
\end{figure}

Table \ref{tab:datasets} summarizes the datasets used in this study and Figure \ref{fig:model} presents the used model architecture.

\begin{table*}
	\centering
	\caption{Overview over the datasets used in this study.}
	\begin{tabular}{ll}
		\toprule
		\bfseries dataset  & \bfseries \# samples  \\
		\midrule
		\multicolumn{2}{c}{Evaluation/Supervised training}\\\midrule
		\textit{PTB-XL} \citep{Wagner:2020PTBXL} & 21,837 \\
		\textit{Chapman} \citep{Zheng2020}  & 10,646\\\midrule
		\bottomrule
	\end{tabular}
	\label{tab:datasets}
\end{table*}

\section{Bootstrap Comparison on pathology Level}
\begin{figure*}
	\centering
	\includegraphics[width=\textwidth]{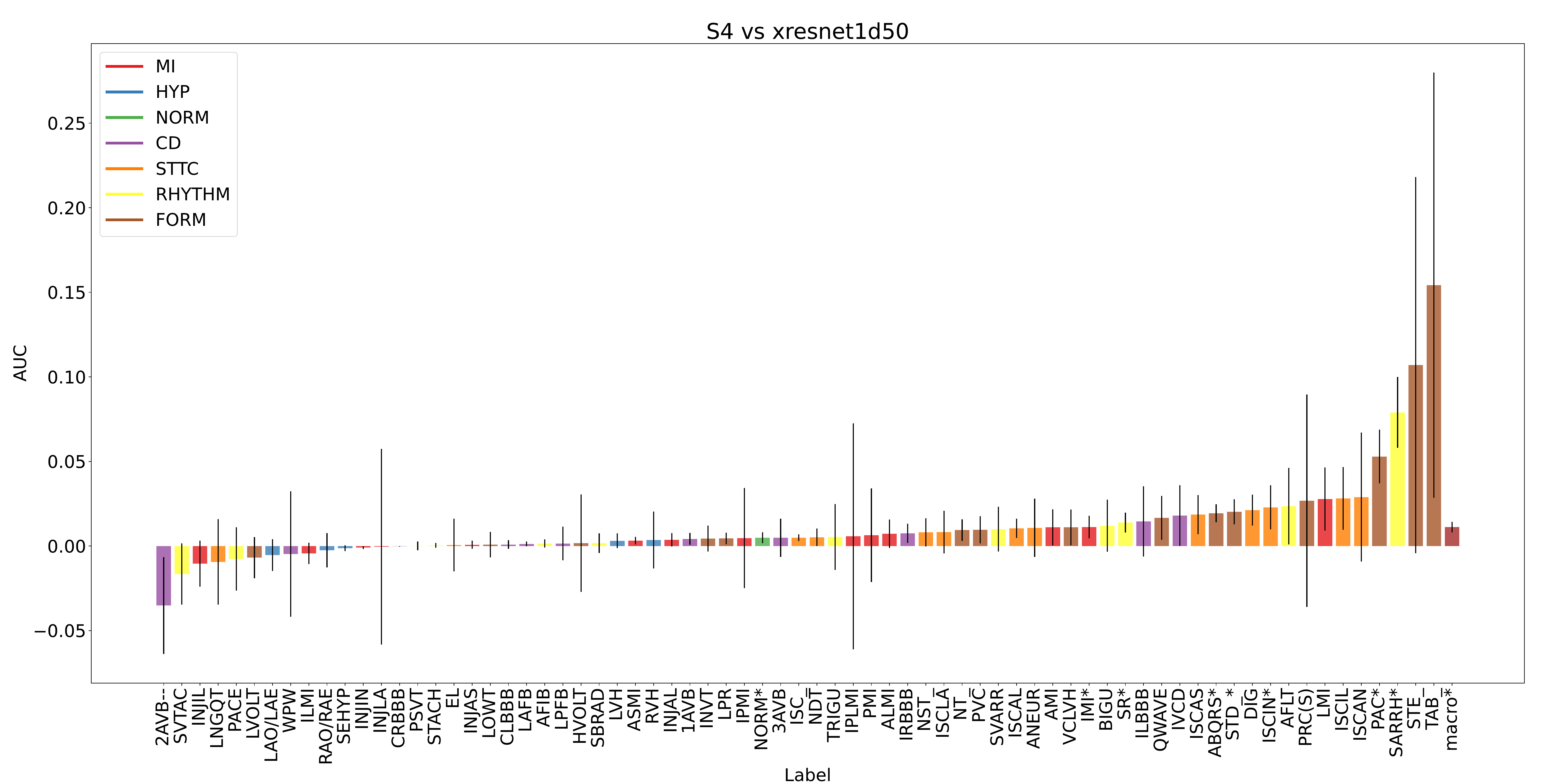}
	\caption{Comparison of \textit{4S4+1FC} and \textit{xresnet1d50} model performance at sampling rate 100Hz. We plot label AUC(4S4+1FC)-label AUC(xresnet1d50). The asterisk means that the \textit{4S4+1FC} model performs statistically significantly better and -- that it performs statistically significantly worse than the \textit{xresnet1d50} model on the respective label.}
	\label{fig:stat_diffs}
\end{figure*}
\label{app:paths}
In Figure \ref{fig:stat_diffs}, we show a bootstrap comparison between the \textit{4S4+1FC} architecture and the \textit{xresnet1d50} architecture on the level of individual ECG statements. The colored bars represent the median values and the black bars the standard deviation of the $n_\text{runs}^2$ medians of the $n_\text{iter}$ AUC difference comparisons per model combination. Labels on which the \textit{4S4+1FC} architecture performs statistically better(worse) are marked by  *(--). The plot reveals that despite high median values for the difference of some label AUCs, like e.g for non-specific ST Elevation (STE) or T-wave abnormality (TAB), the \textit{4S4+1FC} architecture does not necessarily perform statistically better on those labels, as the difference varies strongly over combinations of different runs. On the other hand though, there are pathologies, where the median AUC difference is close to zero but with such a low variance, that these differences are statistically significant, as it is the case for e.g.\ healthy ecg signals (NORM) or inferior myocardial infarctions (IMI).




\end{document}